\documentclass[10pt,twocolumn,letterpaper]{article}
\usepackage{csvsimple}

\usepackage{graphicx}
\usepackage{amsmath}
\usepackage{amssymb}
\usepackage{booktabs}
\usepackage{xcolor}
\usepackage{multirow}
\usepackage{graphicx}
\usepackage{longtable}
\usepackage[english]{babel}
\usepackage{amsthm}
\newcommand{\R}{\mathbb R}
\usepackage{float}
\usepackage[frozencache,cachedir=.]{minted}

\usepackage{cite}

\usepackage[pagebackref,breaklinks,colorlinks]{hyperref}

\newcommand\Tstrut{\rule{0pt}{2.6ex}}
\newcommand\Bstrut{\rule[-1ex]{0pt}{0pt}}

\begin{document}

\title{\bf{EM-driven unsupervised learning \\for efficient motion segmentation}}

\author{Etienne Meunier, Anaïs Badoual, and Patrick Bouthemy \\~\\
Inria, Centre Rennes - Bretagne Atlantique, France}


\maketitle

\begin{abstract}

In this paper, we present a CNN-based fully unsupervised method for motion segmentation from optical flow. We assume that the input optical flow can be represented as a piecewise set of parametric motion models, typically, affine or quadratic motion models. The core idea of our work is to leverage the Expectation-Maximization (EM) framework in order to design in a well-founded manner a loss function and a training procedure of our motion segmentation neural network that does not require either ground-truth or manual annotation. However, in contrast to the classical iterative EM, once the network is trained, we can provide a segmentation for any unseen optical flow field in a single inference step and without estimating any motion models. We investigate different loss functions including robust ones and propose a novel efficient data augmentation technique on the optical flow field,  applicable to any network taking optical flow as input. In addition, our method is able by design to segment multiple motions. Our motion segmentation network was tested on four benchmarks, DAVIS2016, SegTrackV2, FBMS59, and MoCA, and performed very well, while being fast at test time.

\end{abstract}

\section{Introduction}
\label{sec:intro}

Motion segmentation is among the main computer vision tasks. Its goal is to divide a frame into motion-related coherent segments. Motion coherence must be understood with respect to a given property expressed by motion features, parametric motion models, or even higher-level motion information. Depending on the formulation of the problem or the need of the application, segments can be layers, i.e., non necessarily connected subsets of points, or regions, i.e., connected segments, forming a partition of the image grid. Motion segmentation is a relevant step for many applications covering  video interpretation, biomedical imaging, robot vision, and autonomous navigation, to name a few.


Motion segmentation is a complex problem that has been investigated for decades, but there are still open questions. Indeed, it combines topology and information aspects in an intricate way. By topology, we mean the partition of the frame constituting the output of the segmentation. By information, the type of features or motion models it relies on. This all results in a chicken-and-egg problem: estimating easily and correctly the involved motion models requires an available partition, getting an accurate and reliable partition implies available motion models driving the segmentation.

As we want to address general-purpose image motion segmentation without anticipating any given application, we cope with {\color{black} optical flow segmentation (OFS)}. Indeed, the optical flow carries all the information related to motion between two successive frames of a video. {\color{black} Segmenting optical flow enables specific computer vision goals, as for instance, segmenting independent moving objects in the scene. Of course, the image motion depends on one hand on the relative motion between scene objects and the camera, and on the other hand on the object depth. As a consequence, any significant difference in depth results in a distinctive image motion. A static object in the foreground of the scene produces a specific flow pattern compared to the static background, and its image motion is supposed to form a separate segment. In the sequel, we call this situation "motion parallax". However, identifying the nature of the flow segments is outside the scope of this paper. It could be achieved with additional steps as done for example in \cite{meunier2021}.}

We assume that the input optical flow can be represented as a piecewise set of parametric motion models, typically affine or quadratic ones, each of them characterizing motion in one segment. Thus, we {\color{black} formulate OFS} as a piecewise linear regression problem, where finding supports (segments) and estimating the motion models are intertwined issues. 

This problem can be addressed with latent variables, which usually imposes an alternate optimisation strategy. The Expectation-Maximization (EM) algorithm is certainly the flagship solution for a statistical approach of this problem \cite{dempster_EM_1977}. Several extensions to the original EM were proposed as the Classification EM (CEM) introduced in \cite{celeux1992}, where emphasis is put on the clustering issue of the problem beyond the mixture model one. However, classical EM relies on handcrafted features, and leads to time-consuming iterative algorithms. On the other hand, deep learning, and more precisely, convolutional neural networks (CNN), have now become the most effective key solution for image and motion segmentation \cite{badrinarayanan2017,He-mask-RCNN2017,ranjan_competitive_2019,ronneberger_Unet_2015}.
Nevertheless, the training step remains an important issue. Supervised learning provides high accuracy, but manual annotation of optical-flow segmentation maps as ground truth is very cumbersome, and almost unreachable at a large scale. {\color{black} One solution to overcome this issue is to train on synthetic data whose segmentation ground truth is known. However, for some application, the simulation of realistic data is very challenging.} Unsupervised training is thus preferable but 
{\color{black} more challenging to optimize, in particular to formulate the appropriate loss leading to the intended outcomes. Moreover,} an unsupervised method is certainly the best way to deal with videos, and specifically optical flows, unseen during the training phase, thus ensuring better generalisation.

In this paper, we aim to bring the two, EM and CNN, together in order to design a principled and efficient unsupervised motion segmentation method. 
By unsupervised, we mean that we do not resort to any ground-truth and manual annotation, both for the training stage in the loss function, for the selection of the optimal trained network model and for the choice of network hyperparameters.
On one hand, the parametric motion models will carry the coherence for each motion segment. However, the key point is to confine their estimation to the training stage. On the other hand, we take EM as the well-founded basis for the design of the loss function and consequently the training stage of the motion segmentation network. In the related framework of mixture density networks \cite{Bishop_MDN_1994}, a neural network is combined with a mixture density model. However, it does not rely on EM.
Once trained, our network segments each flow of the video without any iteration and any motion model estimation. 

Thus, the main purpose of our EM-driven network is to uncover motion coherence. Taken alone, it can segment motion within videos. It can also be incorporated in a larger framework to solve any video segmentation or understanding problem {\color{black} that could benefit from a motion coherence cue. This could be achieved either by using our network as a module inside a bigger pipeline, for motion saliency detection for instance, or by using our proposed loss as a regularization term for the training of a neural network.}

We demonstrate the efficiency and accuracy of our OFS method on the task of segmenting moving objects in videos,
due to the availability of well-known benchmarks for this application.
We evaluate and compare it to existing methods on four datasets: DAVIS2016 \cite{pont-tuset_benchmark_2016}, FBMS59 \cite{ochs2014}, SegTrackV2 \cite{li-SegTrackv2-2013} and MoCA \cite{lamdouar_camouflage_2020}.

The contributions of our work are summarized below:
\begin{itemize}
\item We infer a principled unsupervised CNN-based motion segmentation method from the EM algorithm;
\item We introduce a new data augmentation scheme adapted to optical flow fields;
\item We are able to segment in a fast and non-iterative way multiple motions using optical flow only;
\item Experiments on several challenging datasets show that our method outperforms, both in terms of accuracy and computational efficiency, comparable unsupervised motion segmentation methods {\color{black} for DAVIS2016 and FBMS59, and is the second best for SegTrackV2 and MoCA.}
\end{itemize}

The paper is organized as follows. Section \ref{related-work} describes related work on motion segmentation. In Section \ref{EM-segmentation}, we formulate the motion segmentation problem through the EM framework.
We present in Section \ref{CNN} how we leverage the EM algorithm to design {\color{black} the network loss function, its architecture and the training procedure}. Section \ref{results} reports extensive experiments with comparison to other existing methods on four benchmarks. In Section \ref{discussion}, we discuss several issues related to the main components of our approach including a comparison with the choices made by other methods. Finally, Section \ref{conclusion} provides concluding remarks.

\section{Related work on motion segmentation}
\label{related-work}

In this section, we comment works related to motion segmentation. For the ease of the presentation, we organize it in three parts, even though they may overlap or a given work may fall into several categories. First, we deal with video object segmentation, where the focus is on primary moving objects (usually a single one followed by the camera). The second category targets the detection of independently moving objects in the scene viewed by a mobile camera. The third one is concerned with image motion segmentation in a more general perspective {\color{black} than the two first more specific categories. Our OFS method belongs to the last category.}

\subsection{Video object segmentation}

The focus of video object segmentation (VOS) is on segmenting primary objects (typically, a single one) moving in the foreground of a scene and usually followed by the camera. VOS delivers a binary segmentation, primary object versus background \cite{wang_survey-VOS}. Nevertheless, it may occur that the background contains moving objects as well, as in some videos of the DAVIS2016 dataset \cite{pont-tuset_benchmark_2016}. The availability of large annotated VOS datasets makes the use of supervised deep-learning techniques possible for VOS. Using jointly object appearance and motion improves performance in VOS as demonstrated for example in \cite{cheng2017} with a two-branch segmentation network, or in \cite{dave2019} with a learning-based spatiotemporal grouping method. In \cite{song_pyramid_2018}, one convolutional and one recurrent network are jointly trained to segment moving objects. A close formulation is proposed in \cite{jain_fusionseg_2017} with a two-stream fully convolutional neural network combining an appearance module with an optical-flow module. MATNet \cite{matnet} also combines motion with appearance using an attention-based architecture allowing better interaction between those two modalities. COSNet \cite{cosnet} trained in a supervised way a co-segmentation module to segment the object common to a pair of frames. Then, they use this module to segment the primary moving object in a sequence.

Unsupervised VOS methods have also been developed. The one described in \cite{papazoglou_fast_2013} exploits image motion boundaries and appearance models to recognize moving objects throughout videos. In \cite{griffin_tukey-inspired_2019}, the authors assume that the moving object has distinctive low-level appearance and motion features, (i.e., orientation and magnitude of flow vectors), compared to the background. They use a Tukey-inspired measure to detect outlier pixels in the images, and label them as belonging to moving objects. In \cite{arp}, the authors exploit the recurrence property of the primary moving object to segment it from the sequence of images. In \cite{yang_unsupervised_2019}, the authors set up an adversarial framework between a generator network producing a hiding mask on the optical flow, and an inpainter network trying to inpaint the flow inside the mask. The rationale is that independent motion cannot be predicted by the surrounding motion. However, this method might be also sensitive to static objects in the foreground generating motion parallax as shown in \cite{meunier2021}. The authors of \cite{dystab} make the dynamic and static models mutually reinforce during the unsupervised training phase, so that the network can precisely detect the objects of interest in the processed images. 

In a different setting, the method described in \cite{mahendran2018} uses optical flow to confirm the validity of a spatial segmentation by assessing that the collective motion of the pixels is coherent in the regions segmented by the network. As we do, they estimate parametric motion models using an external optimisation technique in the network forward pass. However, they do not leverage an EM-based coherence loss as done in our method, and their work relies on a reconstruction loss.
Consequently, their framework is bounded to use Ordinary Least Square to keep the loss differentiable. {\color{black} It cannot involve robust loss functions in contrast to our method.} As a consequence, their method can be disturbed by systematic noise in the input optical flow.

\subsection{Segmentation of independently moving objects}

When the camera is moving, all points in the image exhibit apparent motion. A frequent goal is to segment areas corresponding to objects really moving in the viewed scene, also designated as independently moving objects, or shortly, as independent motions. The output can be either a binary segmentation as in VOS, but this time all independently moving objects on one side and the static background on the other, or, less often, a multi-label segmentation, each moving object being identified by a different label.

A first approach is to cancel the dominant motion in the image generally due to the camera motion. A classical way to compute the dominant motion is to estimate a parametric motion model, affine or quadratic, with a robust function \cite{odobez_robust_1995}. However, a single model cannot usually encompass an entire static scene with objects at different depths. Such a scene configuration raises the motion parallax issue: distinctive segments in the optical flow corresponding to static objects in the foreground.
Different alternatives have been investigated to solve this problem: a stratification of the moving object-detection problem into scenarios from 2D to 3D based on geometrical cues \cite{irani1998}, projective geometry criteria to distinguish motion segments generated by independently moving objects from those induced by static objects in the scene foreground \cite{csurka1999}, multi-frame monocular epipolar constraint of the camera motion \cite{dey2012}, the flow angle likelihood and 3D rigid motion models in \cite{bideau-moving-2016}, long-term analysis by classifying trajectories as background or foreground \cite{wehrwein_video_2017}. In \cite{narayana_complex_2013}, the authors circumvent the problem for a translating camera by using the orientation of the flow vectors.

A supervised learning method is defined in \cite{tokmakov_learning_2017} to infer relevant motion patterns and consequently identify independently moving objects. It is based on a fully convolutional network trained with synthetic video sequences along with their ground-truth optical flow and motion segmentation maps.
Another approach is to compute the so-called static scene flow, that is, the image motion of the whole static scene, induced by the camera motion. Then, independently moving objects can be identified against this static scene flow as in \cite{ranjan_competitive_2019}, where a competitive collaboration between several networks is designed. However, it requires additional ingredients, including the availability of the camera intrinsic parameters, the accurate estimation of the scene depth and of the camera pose. The authors in \cite{bideau_moa-net_2018} only compensate the rotational component of the camera motion, but still need the camera intrinsic parameters.

\subsection{Segmentation of image motion}

A broader perspective is to partition the 2D motion between two successive frames of the video, whatever the source of every individual image motion. It is then a ``pure'' multi-label segmentation problem. Seminal works on image motion segmentation into layers \cite{wang_representing_1994,ayer1995} or into connected regions \cite{PB-EF1993,odobez_mrf-based_1995}, take two successive images as input and estimate a polynomial motion model (typically, affine models) per layer or per region. These methods are respectively based on a clustering framework \cite{wang_representing_1994}, on MDL encoding \cite{ayer1995}, on Markov Random Fields (MRF) with least-square \cite{PB-EF1993} or robust estimation of the parametric motion models \cite{odobez_mrf-based_1995}.

Subsequently, other paradigms were investigated. Let us quote the estimation of a non-parametric mixture with a variant of the EM algorithm and the use of Green’s functions \cite{weiss1997}, a multi-frame approach based on graph cuts with occlusion detection \cite{xiao-shah2005}, a continuous minimization of a single functional involving implicit multiphase level set implementation and favoring motion boundaries of minimal length \cite{cremers2005}, a level set formulation and motion representation with several basis functions \cite{vazquez2006}, the introduction of depth-ordered MRFs and a graph-cut optimisation method over several frames \cite{sun2012}, the use of large time windows, point trajectories and spectral clustering \cite{ochs2014}, and the simultaneous handling of many tracks along with statistics from the appearance model and temporal consistency \cite{li-SegTrackv2-2013}.

The advent of deep learning and the availability of efficient optical flow methods have recently led to new categories of methods. These methods take as input directly the optical flow computed between two successive images, and train neural networks to produce a segmentation. In \cite{yang_motion-grouping_2021}, the authors leverage the slot attention mechanism introduced in \cite{locatello2020}. The designed network handles the two-mask segmentation, and comprises several components, feature encoding, iterative binding, decoding to layers, and flow reconstruction. Besides, the loss function involves an entropy term to make masks as binary as possible, and a temporal consistency term. In \cite{xu2021}, the motion segmentation problem is handled as a multi-type subspace clustering problem by learning nonlinear subspace filters with stacked multi-layer perceptrons. Then, for inference, the authors apply K-means to the output embeddings.

In \cite{lamdouar_bmvc_2021}, the segmentation of moving objects is extended to the recovery of the whole object (i.e., so-called amodal segmentation) even in case of partial occlusion or temporary static state. The authors resort to multi-frame analysis and transformer encoder. If no manual annotation is required, 
{\color{black} the method however benefits from ground-truth segmentation information in the loss function through simulated data mimicking real situations.}


\section{Motion segmentation as an EM problem}
\label{EM-segmentation}
The core idea of our work is to leverage the Expectation-Maximisation framework to design in a well-founded manner the loss function and the training procedure of our OFS neural network. In this section, we first describe one way to use the EM algorithm for optical flow segmentation. 

Optical flow $f \in \mathbb R^{2 \times W \times H}$ is a vector field defined over an image grid $\Omega$ of size $I = W \times H$. We denote by $f_i \in \mathbb R^2$ the motion vector associated to each site $i \in \Omega$ of this grid. We make the assumption that any optical flow field can be decomposed in a set of $K$ segments or layers, each one grouping a (possibly non connected) part of the image grid and exhibiting a coherent motion. In order to enforce coherence, we choose to represent the motion field within each segment $k$ with a parametric model defined by parameters $\theta_k$. We denote $\theta = \{\theta_k, k=1,\ldots,K\}$. In practice, we use polynomial motion models, typically affine (first-degree polynomial) or quadratic (second-degree polynomial) models. Their interest lies in both an easy physical interpretation and an efficient estimation. For instance, a specific 8-parameter quadratic motion model corresponds to the projection, into the image plane, of the rigid motion of a 3D planar surface. {\color{black} Thus, we can account for any slanted almost planar surface that exhibits a smooth depth variation, not only fronto-parallel ones. By almost planar, we mean a negligible depth variation of the object surface compared to its distance to the camera.} Our method could nevertheless accommodate other types of parametric models.


We now consider the likelihood of the optical flow field $f$ given the set of parameters $\theta$, denoted by $p(f|\theta)$. In order to make explicit the partition of $f$ into $k$ segments and the associated individual $\theta_k$'s, we introduce latent variables $z_i$ such that $p(z_i = k | f_i, \theta_k)$ represents the probability that site $i$ belongs to layer $k$. Assuming conditional independence, the logarithm of the likelihood can be written as below in step 1 of eq.\eqref{EM}. Then, we introduce the $z_i$ variables (steps 2 and 3 of eq.\eqref{EM}), and we finally straightforwardly make any positive distribution $q$ appear (step 4 of eq.\eqref{EM}). We have:
\begin{align}
\log(p(f|\theta)) &= \log \prod_i p(f_i | \theta) \nonumber \\
&= \log\prod_i\sum_k p(f_i, z_i^k | \theta_k) \nonumber \\
&= \sum_i\log \sum_k p(f_i, z_i^k | \theta_k) \nonumber \\
&= \sum_i\log \sum_k q(z_i^k) \frac{p(f_i, z_i^k | \theta_k)}{q(z_i^k)},
\label{EM}
\end{align}
 where $z_i^k \triangleq [z_i = k]$. Maximizing $\log(p(f|\theta))$ w.r.t. $\theta$ is obviously complicated, even if it boils down to $k$ maximizations w.r.t. the $\theta_k$'s. Indeed, variables $z_i$ are hidden. To maximize eq.\eqref{EM} w.r.t. $\theta_k$, variables $z_i$ must be available. Accordingly, we need to maximize also w.r.t. the $z_i$'s.

However, we can use the Jensen's inequality ($h(\mathbb E[x]) \geq \mathbb E[h(x)]$ for any concave function $h$), as done in classical EM \cite{murphy_book2012}, in order to build a lower bound $ll(\theta)$ of the log-likelihood $\log(p(f|\theta))$. We get $\log(p(f|\theta)) \geq ll(\theta)$ with:
\begin{align}
\label{eq:eq2}
ll(\theta)
&=\sum_i\sum_k q(z_i^k) \log \frac{p(f_i, z_i^k | \theta_k)}{q(z_i^k)} \nonumber\\
&=\sum_i \sum_k  q(z_i^k)  \log p(f_i, z_i^k | \theta_k) \nonumber \\ &- \sum_i \sum_k  q(z_i^k) \log q(z_i^k),
\end{align}
where the first term of eq.\eqref{eq:eq2} is the expectation over $q(z_i)$ of $\log p(f_i, z_i | \theta)$ and the second term is the entropy that we note $\mathcal H$. The resulting expression of the lower bound is: 
\begin{equation}
\label{eq-classicalEM}
ll(\theta)= \sum_i \mathbb E_{q(z_i)}[\log p(f_i , z_i | \theta) ] + \sum_i \mathcal H(q(z_i)).
\end{equation}
In the classical EM algorithm, one usually takes $q(z_i^k) \triangleq p(z_i^k|f_i, \theta_k)$, i.e., the posterior distribution. 
Then, one alternates between an expectation step where $q(z_i^k), \forall i, k$, is estimated, and a maximization step where $ll(\theta)$ is maximized w.r.t. the $\theta_k$'s. This procedure monotonically increases the log-likelihood until it reaches a local optimum \cite{murphy_book2012}. 

\section{CNN-based motion segmentation}
\label{CNN}

\begin{figure*}[tbh!]
\centering
\includegraphics[width=0.99\linewidth,trim={0cm 3.5cm 0cm 3.5cm},clip]{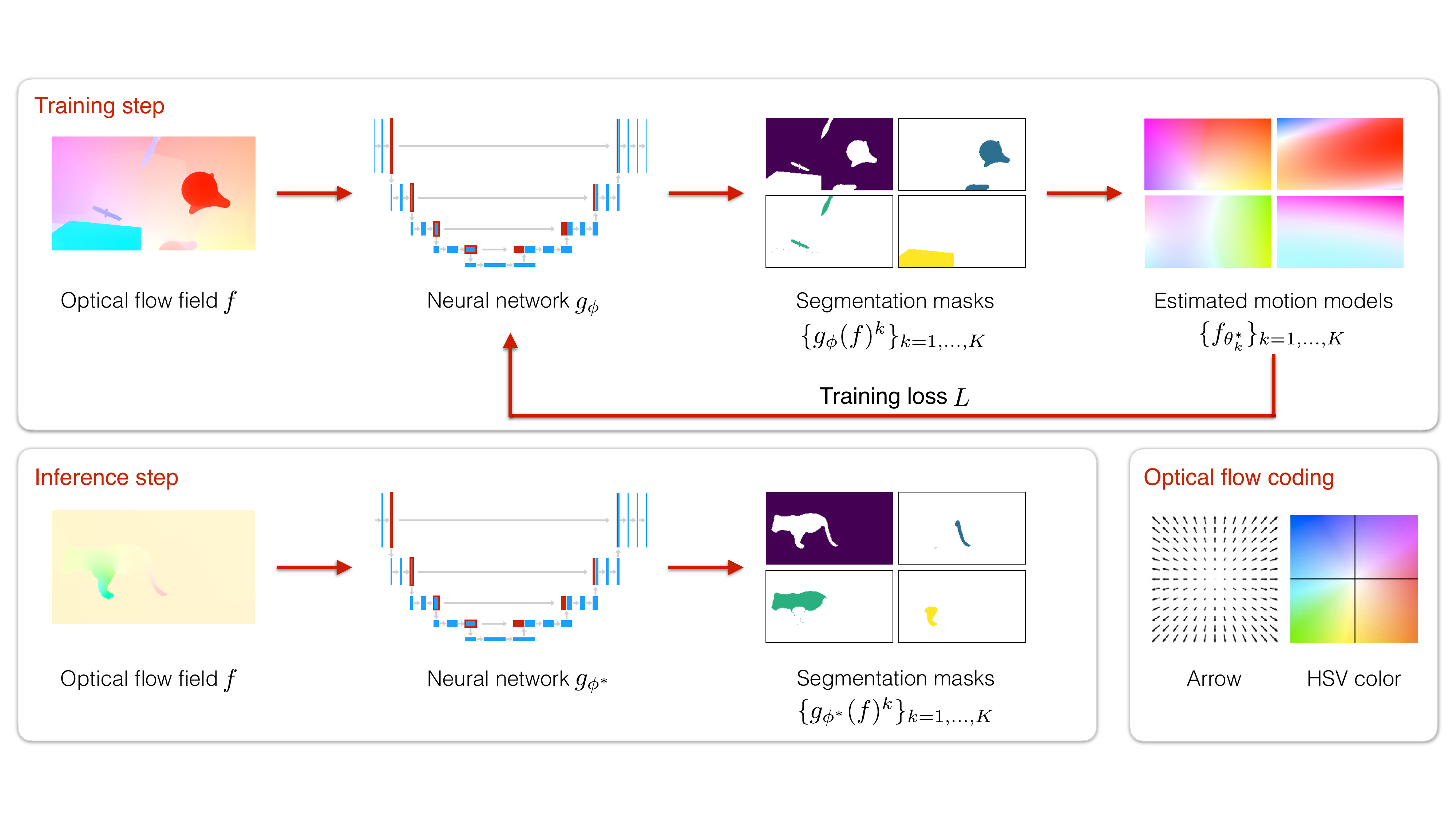}
\caption{Flowchart of the proposed CNN method for the training (top) and inference (bottom) steps. \textit{Training step}: First, we segment the optical flow field $f$ with the neural network $g_\phi$. Then, we get the optimal parametric motion models $\{f_{\theta_k^*}\}_{k=1,\dots,K}$ within each probabilistic segmentation masks $\{g_\phi(f)^k\}_{k=1,\dots,K}$ using~\eqref{eq:theta_optimal}. Finally, we update the parameters $\phi$ of the neural network using~\eqref{eq:phi_optimal}, where the loss function is defined in~\eqref{eq:loss}. This training step is performed iteratively over each batch $\mathcal{B}$ (of size 1 in this illustration). \textit{Inference step}: We directly apply the trained network $g_{\phi^*}$ to any new unseen optical flow field $f$ to obtain the probabilistic segmentation masks $\{g_{\phi^*}(f)^k\}_{k=1,\dots,K}$. There is no estimation of the motion models $\{f_{\theta_k^*}\}_{k=1,\dots,K}$ in the inference step in contrast to the training step. For the sake of visualization, optical flows and polynomial motion models are represented with the HSV color code, but actually, the flow field $f$ used as input of the neural network is taken as a 2D vector field. We have a two-channel input. {\color{black} \textit{Optical flow coding:} correspondence between the arrow visualization of the optical flow field and the HSV color map.}}
\label{fig:flowchart}
 \vspace{-0.3cm}
\end{figure*}

In our case, we adopt a neural network model $g_\phi(f)$, taking as input the optical flow $f$ and parameterized by $\phi$, to produce the image motion segmentation that is our primary goal. {\color{black} This network has a softmax activation in order to output a valid probability distribution over layers for each site.}
The motivation is that, by doing so, we can access a large family of distributions. Most importantly, after the training stage, our network is able to infer the motion segmentation without iterating and without involving motion models,
in contrast to the classical EM algorithm. In addition, it is much faster. At the training stage, we deal with two sets of parameters: the parameters of the motion models $\theta$ and the parameters of the network model $\phi$. At the inference stage, we are only concerned by the network parameters $\phi$.
The overall flowchart of our method, including training and inference stages, is given in Fig.\ref{fig:flowchart}.

\subsection{EM-driven network specification}
\label{sec:EMNet}

Coming back to eq.\eqref{eq:eq2} and following the choice expressed above, we take $g_\phi(f)_i^k$ as $q(z_i^k)$, where $g_\phi(f)_i^k$ is the probability (prediction) given by the network for site $i$ to belong to segment $k$ given the input optical flow $f$.
The lower bound now depends on two sets of parameters, $\theta$ and $\phi$, and writes:
\begin{align}
\label{eq:eq4}
ll(\theta, \phi)
&= \sum_i \sum_k  g_\phi(f)_i^k  (\log p(f_i, z_i^k | \theta_k) - \log g_\phi(f)_i^k)  \nonumber\\
&= \sum_i \mathbb E_{g_\phi(f)_i}[\log p(f_i , z_i | \theta) ] + \sum_i \mathcal H(g_\phi(f)_i).
\end{align}
We alternatively optimize $ll(\theta, \phi)$ with respect to $\theta$ and $\phi$ for the training stage as follows:
\begin{align}
\label{eq:optimtheta}
&\theta^* = \arg \max_{\theta} \sum_i \sum_k g_\phi(f)^k_i \log(p(f_i , z_i^k | \theta_k)) \\
\label{eq:optimphi}
&\phi^* = \arg \max_{\phi}\sum_i \sum_k g_{\phi}(f)_i^k \log(p(f_i , z^k_i | \theta_k^*)) \nonumber\\
&+ \sum_i \mathcal H(g_{\phi}(f)_i).
\end{align}
As previously described in \cite{hathaway1986}, the entropy of the predicted segmentation at each site $i$, $\mathcal{H}(g_{\phi}(f)_i)$, naturally arises in eq.\eqref{eq:eq4}, and then, in eq.\eqref{eq:optimphi}. Entropy measures statistical uncertainty and is maximised for $g_\phi(f)_i^k = \frac 1K, \forall i,k$. It acts as a regularization term balancing the likelihood term to avoid falling too quickly into (inappropriate) local optima. 

Regarding the optimisation on $\theta$, we can reach a local optimum using an off-the-shelf iterative algorithm. However, we can only perform a gradient descent step for the optimisation with respect to the network weights $\phi$.

In order to gain intuition on how the network is learning to produce the motion segmentation, following \cite{murphy_book2012}, we rewrite the lower bound as:
\begin{align}
ll(\theta, \phi) &= \sum_i\sum_k g_\phi(f)_i^k \log \frac{p(f_i, z_i^k | \theta_k)}{g_\phi(f)_i^k} \nonumber\\
&=\sum_i\sum_k g_\phi(f)_i^k \log \frac{p(z_i^k | f_i, \theta_k)}{g_\phi(f)_i^k} \nonumber\\
&+ \sum_i \log p(f_i | \theta) \sum_k g_\phi(f)_i^k \nonumber\\
&= - \sum_i \mathbb{KL}[g_\phi(f)_i || p(z_i|f_i, \theta)] + \log(p(f|\theta)).
\end{align}
Consequently, the optimisation step over the network weights is defined by: 
\begin{equation}
\phi^* = \arg \min_{\phi} \sum_i \mathbb{KL}[g_\phi(f)_i || p(z_i|f_i, \theta^*)],
\end{equation}
where we minimize the KL-divergence between the segmentation produced by the network and the segmentation linked to the optimal parameters $\theta^*$. Thus, the network is trained to produce a segmentation for a given set of parameters. As the quality of the network segmentation improves, so does the quality of the estimated $\theta^*$, pushing the network weights to produce better segmentation in turn.


\subsection{Flow likelihood and loss function}
\label{sec:dataterm}

In the previous section, we described the overall training process. In this section, we address the definition of the different terms of the loss function. 

First, we decompose the joint probability in eq.\eqref{eq:eq4} into a likelihood and a prior:
\begin{equation}
\label{eq:eq9}
    p(f_i, z_i^k| \theta_k) = p(f_i | z_i^k, \theta_k)p(z_i^k).
\end{equation}
The likelihood $p(f_i | z_i^k, \theta_k)$ assesses how the estimated parametric motion model in a given region fits the observed flow in this region. In this work, we use a uniform prior for $p(z_i^k)$. Nevertheless, we could adopt a more complex prior, if we wanted to influence the size of each region for instance.

An important point of our design is to specify the form of the likelihood $p(f_i | z_i^k, \theta_k)$ that is used to compare the input optical flow with the parametric flow for a given set of parameters $\theta$. In practice, since our parametric motion models are dependent on the position of the points on the 2D space, we introduce a deterministic function $c(i)$ that maps the site $i$ to a polynomial expansion involving its coordinates. More specifically, for a 6-parameter affine motion model, we have $c(i) = \begin{bmatrix}1, x_i, y_i\end{bmatrix}$; for a full 12-parameter quadratic model, we have $c(i) = \begin{bmatrix}1, x_i, y_i, x_i^2, x_iy_i, y_i^2\end{bmatrix}$. The likelihood evaluates the distance between the input flow vectors $f$ and the parametric flow vectors $f_{{\theta_k}_i} \triangleq  \theta^T_k \cdot c(i), \forall k,i$. Its general form is given by:
\begin{equation}
\label{eq:likelihood}
p(f_i |z_i^k, \theta_k) = \frac1{Z} \exp(-{\color{black}\frac{1}{\alpha}}\delta(f_i, \theta^T_k \cdot c(i))),
\end{equation}
where $\delta :\mathbb{R}^{2*2} \to \mathbb{R}$ is a distance function to define, {\color{black} and $\alpha$ is a free parameter related to the uncertainty in the flow measure and the resulting adequacy of the parametric motion model.} If $\delta$ is a translationally invariant function, which we verified for all tested distance functions, then $Z$ is only dependent on the function $\delta$ {\color{black} and hyperparameter $\alpha$} and not on the input. The proof can be found in the supplementary material. This allows us to perform optimisation without explicitly computing $Z$.

The choice of the distance function $\delta$ is central in our approach, as it is used both for the estimation of the parametric motion models and for the training of the network (see eq.\eqref{eq:optimtheta} and eq.\eqref{eq:optimphi}). Robust loss functions can be beneficial as thoroughly investigated in \cite{barron2019}. We consider the following distance functions: 
\begin{itemize}
\item Squared $L_2$ : $\delta(f_i, \theta^T_k \cdot c(i)) = 
||f_i -\theta^T_k \cdot c(i)||_2^2$
\item  $L_2$ norm : $\delta(f_i, \theta^T_k \cdot c(i)) = 
||f_i -\theta^T_k \cdot c(i)||_2$
\item  $L_1$ norm : $\delta(f_i, \theta^T_k \cdot c(i)) = 
||f_i -\theta^T_k \cdot c(i)||_1$
\end{itemize}
$L_1$ (due to the absolute function involved) and $L_2$ (due to the square root of the sum involved) norms bring robustness to outliers in the flow field, in contrast to the squared $L_2$.

We define the loss of our model as:
\begin{equation}
L(f, \theta, \phi) = - ll(\theta, \phi),
\end{equation}
where $ll(\theta, \phi)$ is given by eq.\eqref{eq:eq4}. Taking into account eq.\eqref{eq:likelihood}, we can formulate the loss function as: 
\begin{align}\label{eq:loss}
L(f, \theta, \phi) &= \frac{1}{\alpha}\sum_i \sum_k  g_\phi(f)_i^k  \delta(f_i, \theta_k^T \cdot c(i)) \nonumber \\ &+ \sum_i \sum_k  g_\phi(f)_i^k \log g_\phi(f)_i^k + I\  \log(K \ Z),
\end{align}
where {\color{black} $Z$ is the normalization term in eq.\eqref{eq:likelihood} and} $\alpha$ 
allows us to balance the likelihood, prior and entropy parts of the loss.
We use $\alpha=10^{-2}$ in all our experiments, but in practice the network model is fairly robust to the choice of this hyperparameter. {\color{black}In the supplementary material, more details are given on the derivation of eq.\eqref{eq:loss}.} 

\subsection{Network training}
\label{training}

During the training, we minimize the loss function $L$ over a dataset of optical flow fields. For each input flow field $f$ of the training dataset, we minimize $L(f, \theta, \phi)$ with respect to each parameter. This alternate optimisation is performed over every batch $\mathcal{B}$ as follows: 
\begin{align}\label{eq:theta_optimal}
\theta^* &= \arg \min_{\theta} \sum_{f \in  \mathcal{B}} L(f, \theta, \phi^{t})
\end{align}
\begin{align}\label{eq:phi_optimal}
\phi^{t+1} &= \phi^t - \gamma \nabla_{\phi}  \sum_{f \in  \mathcal{B}} L(f, \theta^*, \phi),
\end{align}
where $t$ is the iteration number and $\gamma$ the learning rate. 

\begin{figure}[h!]
\includegraphics[width=8cm]{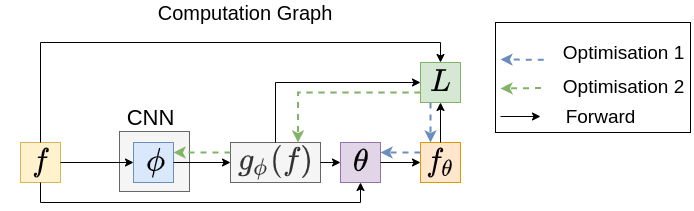}
\caption{Illustration of the computation graph for the training of our network. $m \triangleq g_{\phi}(f)$ denotes the set of arrays (as many as masks) collecting the probability for each site of the input flow field to belong to each mask. $\theta$ is the set of the motion model parameters, and $\phi$ the set of the network parameters. In our method, we are alternatively optimising w.r.t. $\theta$ (optimisation 1) and $\phi$ (optimisation 2).}
\label{fig:computationgraph}
 \vspace{-0.1cm}
\end{figure}

In practice, we use an optimizer to get $\theta^*$ and an automatic differentiation to compute the gradients with respect to $\phi$. As described in subsection \ref{sec:EMNet} and in our computation graph presented in Fig.\ref{fig:computationgraph}, we consider $\theta^*$ as fixed in the gradient step with respect to $\phi$, making $\nabla_{\phi} L(f, \theta^*, \phi)$ trivial to compute using automatic differentiation. Practical details are provided in subsection \ref{implementationdetails} 

\subsection{Data augmentation}
\label{augmentation}
As proven beneficial in many computer vision problems, we proceed to data augmentation to train the motion segmentation network. However, the input data are not images but optical flows in our case, which led us to define an original data augmentation procedure. The goal is to make the network as invariant as possible to the global motion field, since we identify it as an important clue for generalisation.

Therefore, we add to each optical flow field of the dataset a parametric motion model whose parameters are drawn at random. For the sake of consistency, we take quadratic motion models. This mimics a large variety of camera movements. This procedure has the advantage of multiplying the flow configurations, while keeping the same flow structure as in the initial sample. In other words, the network is trained with a diversity of flows, while having the same target segmentation to predict. In the supplementary material, we give a formal proof that our loss function is invariant to the added parametric global motion. As shown in Section \ref{ablation}, this data augmentation scheme contributes to improve the overall performance of the network. Let us emphasize that it could also be used to train any network taking optical flow as input with the benefit of such invariance.




\begin{table*}[tb!]
\centering{
    \resizebox{\linewidth}{!}{\begin{tabular}{|l || l|l|c c|c|c|c|}
    \hline
        Method & Training & Input & \multicolumn{2}{c|}{DAVIS2016} & SegTrack V2 & FBMS59 & MoCA \Tstrut\\
        ~ & ~ & ~ & $\mathcal{J}$ &  $\mathcal{F}$ &  $\mathcal{J}$ &  $\mathcal{J}$ &  $\mathcal{J}$ \Bstrut\\ \hline
        \bf{Ours} & \multirow{7}{*}{Fully\ Unsupervised} & \multirow{4}{*}{Flow} & 69.3 & 70.7 & 55.5 & 57.8 & 61.8 \Tstrut\Bstrut\\ 
        \cline{4-8}
        MoSeg \cite{yang_motion-grouping_2021} &&  & 68.3 & 66.1 & 58.6 & 53.1 & 63.4 \Tstrut\Bstrut\\  \cline{4-8}
        FTS \cite{papazoglou_fast_2013} &&  & 55.8 & ~ & 47.8 & 47.7 & - \Tstrut\Bstrut\\ \cline{4-8}
        TIS$_0$ \cite{griffin_tukey-inspired_2019} &&  & 56.2 & 45.6 & - & - & - \Tstrut\Bstrut\\   \cline{1-1} \cline{3-8}
        TIS$_s$ \cite{griffin_tukey-inspired_2019} && \multirow{3}{*}{Image \& Flow}  & 62.6 & 59.6 & - & - & - \Tstrut\Bstrut\\  \cline{4-8}
        CIS - No Post \cite{yang_unsupervised_2019} & & & 59.2 & ~ & 45.6 & 36.8 & 49.4 \Tstrut\Bstrut\\ \cline{4-8}
        CIS - With Post \cite{yang_unsupervised_2019} && & 71.5 & ~ & 62.0 & 63.6 & 54.1 \Tstrut\Bstrut\\ \hline \hline
        DyStab - Dyn \cite{dystab} & \multirow{3}{*}{Supervised Features}& Flow & 62.4 & ~ & 40.0 & 49.1 & - \Tstrut\Bstrut\\ \cline{3-8}
        DyStab - Stat\&Dyn \cite{dystab} & & Image \& Flow & 80.0 & ~ & 73.2 & 74.2 & - \Tstrut\Bstrut\\ \cline{3-8}
        ARP \cite{arp} && Image \& Flow & 76.2 & 70.6 & 57.2 & 59.8 & - \Tstrut\Bstrut\\ \hline \hline
        MATNet \cite{matnet} & \multirow{2}{*}{Supervised} & Flow & 82.4 & 80.7 & ~ & ~ & 64.2 \Tstrut\Bstrut\\\cline{3-8}
        COSNet \cite{cosnet} && Image & 80.5 & 79.5 & - & 75.6 & 50.7 \Tstrut\Bstrut\\ \hline
    \end{tabular}}}
    \vspace{0.3cm}
 \caption{Results on DAVIS2016 validation dataset, SegTrackV2, FBMS59 validation dataset and MoCA for several unsupervised and supervised methods (scores taken from \cite{griffin_tukey-inspired_2019}, \cite{yang_unsupervised_2019}, \cite{yang_motion-grouping_2021} and \cite{dystab}).
 $\mathcal{J}$ is the Jaccard index (region similarity) and $\mathcal{F}$ accounts for contour accuracy. The higher the value, the better the performance. For further explanation on the evaluation metrics, we refer the reader to the DAVIS2016 website. Following the evaluation protocols, reported scores are the average of scores over all samples in the corresponding dataset, except for DAVIS2016 where it is the average of each sequence average score. {\color{black} Our network is trained once and for all on the synthetic dataset FT3D whatever the benchmark.}
 }
 \label{tab:results}
 \vspace{-0.6cm}
\end{table*}

\subsection{Related work on EM and deep learning}

To conclude this section, we explain how our approach differs from other works that somehow make use of EM within a neural network framework. 
In \cite{li2019}, the EM paradigm is involved in the network designed for semantic segmentation. However, the purpose is quite different since they exploit the EM algorithm in the attention mechanism. It allows them to iteratively estimate a compact set of bases used to compute the attention maps.

The EM framework plays also a role in \cite{greff2016,greff2017}, and just recently in \cite{Yu-2021}, in the design of the neural network architecture concerned with perceptual grouping tasks. However, those approaches differ in several ways. First, they remain iterative. One starts from parameters specifying each latent space component (as our motion parameters $\theta$), but the network iteratively refines them in the inference stage, so that this network remains dependent on the initialization. Another big difference with our method lies in the network architecture itself, which includes generative branches to produce an approximation of the input data from the latent space components and an iterative branch (recurrent network in \cite{greff2016,greff2017} and sampling in \cite{Yu-2021}) to update the latent space components themselves. 
Since those methods rely on an iterative inference process implementing EM, they rather belong to the class of algorithm unfolding techniques \cite{monga2021}. 

\section{Experimental results}
\label{results}

\subsection{Implementation details}
\label{implementationdetails}

Optical flow fields are computed on the original video frames using the RAFT method \cite{teed_raft_2020}. Then, we downsample them to obtain $128 \times 224$ vector fields provided as input to the network. The resulting segmentation is subsequently upsampled to the original frame size for evaluation w.r.t. the ground truth. It allows us to perform much more efficient training and inference stages. In all experiments, we use the $L_1$ norm loss function, unless otherwise specified.

We take the full quadratic motion model with 12 parameters to represent the optical flow within each segment $k$:
\begin{align}
f_{\theta_k}(x,y) &= (\theta_{k_1}+\theta_{k_2}x+\theta_{k_3}y+\theta_{k_4}x^2+\theta_{k_5}xy+\theta_{k_6}y^2,\nonumber\\
& \theta_{k_7}+\theta_{k_8}x+\theta_{k_9}y+\theta_{k_{10}}x^2+\theta_{k_{11}}xy+\theta_{k_{12}}y^2)^T,
\end{align}
where point $(x,y)$ belongs to segment $k$. We take this parametric motion model since it can better fit complex motion. {\color{black} Indeed, it is likely to better encompass} the background motion when the camera motion includes both translation and rotation with a static {\color{black} background involving objects at slightly} different depths, and for articulated motions as well.

Our method is fully unsupervised. We do not resort to any manual annotation for training nor for model selection. Indeed, in all experiments, we selected the stopping epoch from the loss function evaluated always on the same validation set: the official training split of DAVIS2016 \cite{pont-tuset_benchmark_2016}. 

We consider generalisation to unseen datasets as an essential property of any motion segmentation method. To apply this idea, we trained once and for all our model on a single dataset, namely Flying Things 3D (FT3D) \cite{ft3d2016}, for all experiments.
In addition, the use of a synthetic dataset alleviates the problem of choosing the relevant real training data for a given experiment. For completeness, we also performed training experiments on real datasets, and obtained similar or slightly better performance depending on the chosen real training set.

Let us recall that the optimisation on $\theta$ does not occur at inference stage. The network prediction for each site of the flow field to belong to each segment $k$ is directly used to yield the motion segmentation map. We simply select for each site segment $\hat{k}$ with the highest probability. 

No postprocessing is performed on the resulting segmentation. As shown later, the obtained segments are generally smooth, certainly due to the implicit regularization capacity of the network.

We take as input the optical flow in its vector field representation $f\in \mathbb R^{W \times H \times 2}$. Thus, we have a two-channel input for the network. Our loss function and our training procedure could be adapted to any neural network designed for segmentation. We choose the well-known convolutional architecture U-Net \cite{ronneberger_Unet_2015} for $g_{\phi}$. We use a slightly modified implementation of the one available under PyTorch Lightning \cite{lightningbolt20}. We take seven downsampling layers and start with a feature depth of 64. The selection of the structure is again unsupervised using the validation loss on FT3D dataset. As did in \cite{meunier2021}, we use InstanceNorm between convolutional blocks in order 
to tackle variations in optical flow magnitude over the dataset.

We use Adam \cite{adam2014} optimizer with a learning rate of $10^{-4}$ to train the network. 
The optimisation on $\theta$ is done with Pytorch implementation of L-BGFS \cite{lbgfs89}. {\color{black} Batches comprise flow fields randomly sampled from the training dataset.}

Our network\footnote{https://github.com/Etienne-Meunier/EM-Flow-Segmentation} is time efficient, being a simple convolutional network, with an average computation time of $0.008s$ per $128 \times 224$ input flow field on a Tesla-V100, if we consider a batch of size $32$. Without any parallelisation (batch size of $1$), it can run at 36fps making it usable for real-time applications. In particular, it is faster than the fastest method introduced in \cite{yang_motion-grouping_2021}, because we do not use an iterative attention module, thus reducing our computational complexity. In contrast to methods involving self attention, there is a linear relationship between the complexity of our network and the size of the optical flow field, which allows us to readily process input of large dimensions.


\begin{figure*}[tbh]
\includegraphics[width=\linewidth]{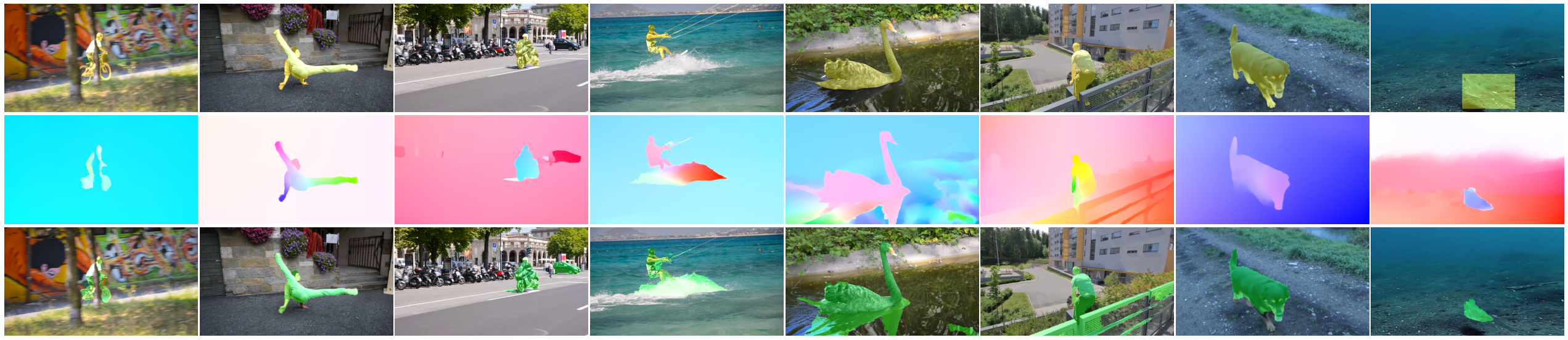}
 \vspace{-0.5cm}
\caption{Examples of motion segmentation results obtained with our method with two masks, on the videos bmx-trees, breakdance-flare, scooter-black, kite-surf, blackswan, parkour, dogs02 and cuttlefish 1, of the DAVIS2016, FBMS and MoCA datasets. First row: a frame of the video with the ground-truth superimposed in yellow. Second row: the input flow field displayed with the HSV color code \cite{middlebury} {\color{black} that is depicted in Fig.\ref{fig:flowchart}}. Third row: the segmentation produced by our method superimposed in green on the corresponding image.}
\label{fig:2masks}
 \vspace{-0.15cm}
\end{figure*}

\subsection{Comparative evaluation}
\label{eval}

We want to objectively evaluate the performance of our method for segmenting optical flow fields. Due to the lack of benchmarks dedicated to OFS, we compare our method on four VOS datasets described below.

\textbf{DAVIS2016}\footnote{https://davischallenge.org/index.html} \cite{pont-tuset_benchmark_2016} includes 50 videos (3455 frames) split in 30 train and 20 validation videos with diverse objects. Only the principal moving object is annotated in the ground truth. We use the official criteria for evaluation on this dataset: the Jaccard score and the contour accuracy score.

\textbf{SegTrackV2}\footnote{https://paperswithcode.com/dataset/segtrack-v2-1} \cite{li-SegTrackv2-2013} and \textbf{FBMS59} \cite{ochs2014} respectively comprise 14 videos (1066 annotated frames) and 59 videos (720 annotated frames), each involving one or several moving objects. For FBMS59 we use the 29 sequences of the validation set for evaluation. In cases where there are several moving objects, we group them into a single foreground mask for evaluation, as done in \cite{yang_motion-grouping_2021}.

\textbf{Moving Camouflaged Animals (MoCA)} \cite{lamdouar_camouflage_2020} presents camouflaged animals in natural scenes. For a fair comparison, we use the subset released by \cite{yang_motion-grouping_2021} with 88 videos and 4803 frames. Ground-truth bounding boxes are provided instead of masks for evaluation. Accordingly, we convert our output to a bounding box around the largest connected region of our output mask as done in \cite{yang_motion-grouping_2021}.

Apart from a few videos in SegTrackV2 and FBMS59, those datasets focus on one primary moving object. Indeed, videos depict one single independently moving object in the foreground. Consequently, the ground truth comprises only two segments: foreground primary moving object versus background. To be coherent with this status, we apply our method with two masks, i.e., $K=2$. To choose the foreground mask, we rely on a simple heuristic where we designate the biggest mask as the background one. 

We compare our method with several other supervised and unsupervised methods: MoSeg \cite{yang_motion-grouping_2021}, TIS (two versions) \cite{griffin_tukey-inspired_2019}, CIS \cite{yang_unsupervised_2019}, FTS \cite{papazoglou_fast_2013}, DyStab \cite{dystab}, ARP \cite{arp}, MATNet \cite{matnet} and COSNet \cite{cosnet}. All these methods were described in Section \ref{related-work}. Results are collected in Table \ref{tab:results}. For a fair comparison, we underline several points in this table. First, we differentiate methods trained on ground-truth segmentation masks such as COSNet \cite{cosnet} or MATnet \cite{matnet}, and unsupervised methods. We also indicate methods that use features trained in a supervised way, as it provides a strong advantage, and then do not fit in an unsupervised scenario. Indeed, DyStab \cite{dystab} resorts to supervised classification features to initialise its networks, and ARP \cite{arp} requires a motion boundary detection algorithm previously trained in a supervised way. Secondly, as we evaluate here motion segmentation based on optical flow, we dissociate methods that take RGB images as input from methods using only optical flow like ours. Some works like \cite{griffin_tukey-inspired_2019, dystab} propose versions with (TIS$_s$, DyStab-Stat\&Dyn) and without (TIS$_0$, DyStab-Dyn) RGB frames taken as input, illustrating the influence of this input modality on the final results. As in \cite{yang_motion-grouping_2021}, we distinguish the CIS version involving a strong CRF-based postprocessing, "CIS-With Post", and one without postprocessing, "CIS-No Post", since postprocessing drastically increases runtime making the method unusable for practical applications. It is measured in \cite{yang_motion-grouping_2021} that CIS \cite{yang_unsupervised_2019} has a runtime of 11s/frame with postprocessing against 0.1s without.


Table \ref{tab:results} shows that our method outperforms all comparable methods on DAVIS2016 and FBMS59. Our method is the second best for the two other datasets. It is even close to the best one and far ahead the other comparable methods for MoCA. By comparable methods, we mean unsupervised methods without unusable postprocessing. Let us stress that we train our method on an external dataset, unlike CIS that is trained on test data as well. Scores obtained by our method for every sequence of the four datasets are collected in the supplementary material. {\color{black} In addition, a quantitative comparison, in term of speed and accuracy, between classical EM and our OFS network is given in Table 1 of the supplementary material, along with an experimental study on the dependency of the classical EM to the initialization of the motion model parameters.}


In Fig.\ref{fig:2masks}, we report visual results to figure out how our method behaves on different typical examples from the VOS datasets. We can observe four examples of failure cases with respect to the DAVIS2016 ground truth (i.e., scooter-black, kite-surf, blackswan and parkour), although the extra parts segmented by our method make sense w.r.t. OFS task. Let us recall that the VOS task takes into account by construction only the primary moving object and not all moving objects in the scene. In Scooter-black, the car segmented in the background is moving; in Parkour, the fence is segmented as it exhibits an important motion parallax; in Kite-surf and Blackswan, the ripples on the water are segmented too. This type of complex examples will be more appropriately handled with the multiple motion segmentation described in subsection \ref{sec:multi_seg}.

\vspace{-0.1cm}
\subsection{Ablation study}
\label{ablation}

In order to identify the contribution of the different components of our method, we performed an ablation study. We investigated the following three main components:
\begin{itemize}
    \item DA: removal of the data augmentation on the optical flow described in subsection \ref{augmentation}.
    \item Quad. Model: replacing the full quadratic motion model with the affine one.
    \item $L_1$ norm (a): substitution of the robust loss function ($L_1$ norm) by its non-robust counterpart, the squared $L_2$ given in subsection \ref{sec:dataterm}.
    \item $L_1$ norm (b): substitution of the robust loss function ($L_1$ norm) by the $L_2$ norm given in subsection \ref{sec:dataterm}.
\end{itemize}
We performed each change on our full method one by one.
As shown in Table \ref{tab:ablation}, the quadratic motion model plays a pivotal role in the performance of the network. The other components still play an important role in an equal measure.
We can observe that the $L_2$ norm provides better performance than the squared $L_2$. As mentioned before, we train the network on the FT3D dataset where estimated optical flow fields exhibit low noise, which mitigates the impact of a robust loss for training. In previous experiments, the performance gap between a model trained with $L_1$ loss and squared $L_2$ loss on a real dataset was even more significant.
%
%
\begin{table}[t!]
\resizebox{\columnwidth}{!}{
\begin{tabular}{|c | c c c c || c|}
 \hline
 \bf{without} & DA &  Quad. Model & $L_1$ norm (a) & $L_1$ norm (b) & full method \Tstrut\Bstrut\\
 \hline\hline
 $\mathcal{J}$ Mean $\uparrow$ & 58.4 & 53.1 & 58.8 & 60.6 & 61.1\Tstrut\Bstrut\\
 \hline
\end{tabular}}
\vspace{0.1cm}
\caption{Ablation study for different components of our method. Each time, we suppress or modify only one component, respectively, removal of the data augmentation, substitution of the quadratic motion model by the affine motion model, substitution (a) of the $L_1$ norm by the squared $L_2$, substitution (b) of the $L_1$ norm by the $L_2$ norm. $\mathcal{J}$ is given as the average of the average frame Jaccard index obtained for each of the four datasets presented in Section \ref{eval}.}
\label{tab:ablation}
\vspace{-0.7cm}
\end{table}

\vspace{-0.4cm}
\subsection{Multiple motion segmentation}
\label{sec:multi_seg}
\begin{table*}[h!]
\centering{
    \begin{tabular}{|c||c|c|c|c|}
    \hline
        Our method with $K$ masks & Davis2016 ($\mathcal{J}$) & SegTrack V2 ($\mathcal{J}$) & FBMS59 ($\mathcal{J}$) & MoCA ($\mathcal{J}$)\Tstrut\Bstrut\\ \hline
        \color{black}\bf{$K=3$} & \color{black}75.1 & \color{black}55.0 & \color{black}62.1 & \color{black}64.4 \Tstrut\Bstrut\\ \hline
        \bf{$K=4$} & 76.0 & 59.6 & 64.7 & 66.8 \Tstrut\Bstrut\\ \hline
        \color{black}\bf{$K=5$} & \color{black}77.2 & \color{black}59.7 & \color{black}65.1 & \color{black}66.6 \Tstrut\Bstrut\\ \hline
        \color{black}\bf{$K=6$} & \color{black}78.3 & \color{black}62.0 & \color{black}66.0 & \color{black}68.0 \Tstrut\Bstrut\\ \hline
        \color{black}\bf{$K=7$} & \color{black}78.3 & \color{black}62.8 & \color{black}66.9 & \color{black}67.5 \Tstrut\Bstrut\\ \hline
    \end{tabular}}
     \vspace{0.2cm}
 \caption{Results on DAVIS2016 validation dataset, SegTrackV2, FBMS59 validation dataset and MoCA {\color{black} obtained with our method for several numbers of masks. Quantitative evaluation is performed here using ground truth for selecting the right masks as described in the main text.}}
 \label{tab:multiple}
 \vspace{-0.5cm}
\end{table*}
\begin{figure*}[]
\includegraphics[width=\linewidth]{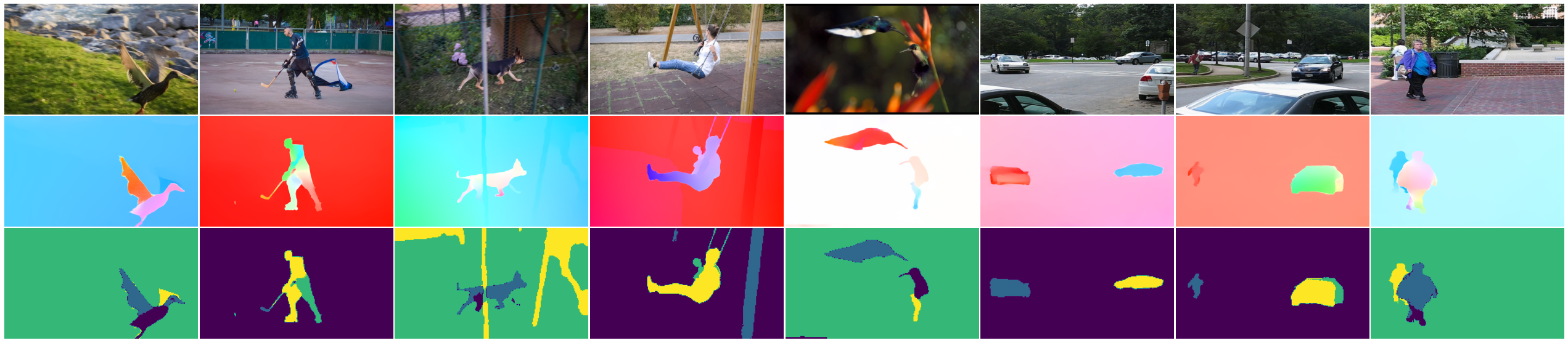}
\vspace{-0.5cm}
\caption{Results obtained with our method for four masks ($K=4$) regarding multiple motion segmentation. First row: one image of the video. Second row: input optical flow displayed with the HSV color code {\color{black} that is depicted in Fig.\ref{fig:flowchart}}. Third row: motion segmentation maps with four masks, one colour per mask (the four masks may be not present if not necessary). We adopt the same color code for all segmentation maps (dark blue: mask 1, light blue: mask 2, green: mask 3, yellow: mask 4). Examples are drawn from DAVIS2016, FBMS59 and SegTrackV2 datasets. Videos in lexicographic order: mallard-fly, hockey, libby, swing, hummingbird, cars5, cars4, people2.}
\label{fig:4masks}
    \vspace{-0.1cm}
\end{figure*}

Our method can handle by design the segmentation of multiple motions. Indeed, it has been defined from the start with $K$ masks. In Section \ref{eval}, we took into account only two masks for the evaluation on VOS datasets, because the challenge and the ground-truth have been defined in this way. In this subsection, we report additional experiments with {\color{black} several masks ($K\in \{3,4,5,6,7\}$) in Table \ref{tab:multiple}. Visual results are also displayed in Fig.\ref{fig:4masks} for $K=4$}. They were obtained on videos from DAVIS2016, FBMS59 and SegTrack-V2. We observe that our method can deal with multiple motions in the video and correctly segment them. This figure includes several examples of articulated motion (e.g., mallard-fly, hockey), and independently moving objects (e.g., cars5, cars4, people2). The figure also contains examples corresponding to ``virtual" failure cases encountered in the two-mask VOS challenge reported in Section \ref{eval}. We mean segments that really correspond to moving objects but are not included in the ground truth that is focused on the primary moving object.
These results demonstrate that, when involving {\color{black} more than two} masks, we can correctly deal with interfering motions such as motion parallax (e.g., libby, swing) or additional moving objects (e.g., cars5). {\color{black} In addition, our multimask parametric motion segmentation could deal with objects comprising several significant planar surfaces of distinct orientation, each of them corresponding then to different optical flow segments.}

{\color{black} We evaluated our multi-mask segmentation on the VOS datasets presented in Section \ref{results}}. As we generate now several masks, segment selection is an issue for computing the $\mathcal{J}$ score that is based on two masks (foreground moving object versus background). Just to be able to compute this score and to figure out the performance we could reach (at least an upper bound), we declare as foreground all layers that overlap for their most part the ground-truth foreground mask. Scores are collected in Table \ref{tab:multiple}. {\color{black} As expected, performance increases with the number of masks up to a certain point.} We observe that we get very high scores on the four benchmarks, {\color{black} and, for $K>3$, we outperform} all the other comparable methods by a relatively large margin (see Table \ref{tab:results}). We are even close to the performance of supervised methods.

Of course, this would not be a valid procedure to extract the primary moving object in practice, since we need the ground-truth masks. However, it gives an insight into the quality of our multiple motion segmentation, since, so doing, we are not modifying our segmentation masks, but only selecting them. It also highlights the potential of our multiple motion segmentation method, if paired with an efficient mask selection procedure.

\begin{figure*}[tbh!]
\includegraphics[width=\linewidth]{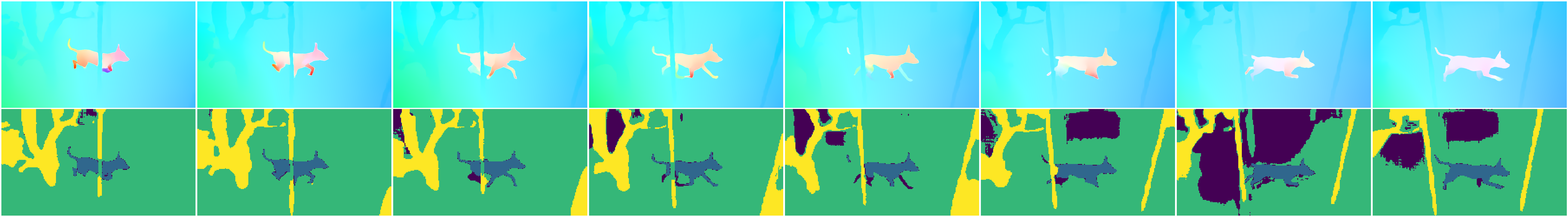}
\includegraphics[width=\linewidth]{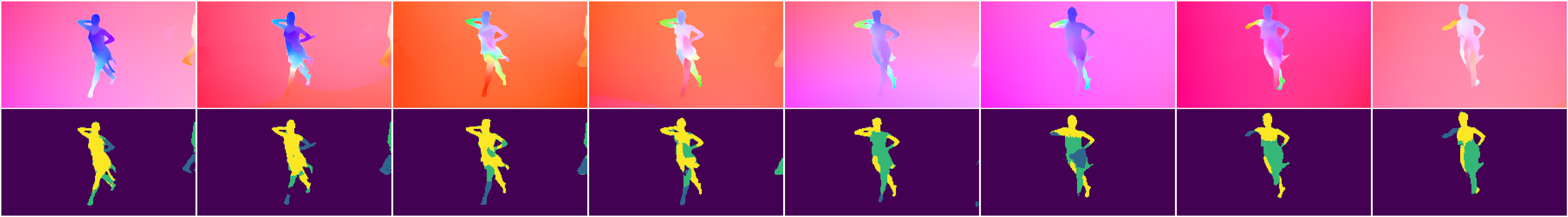}
\includegraphics[width=\linewidth]{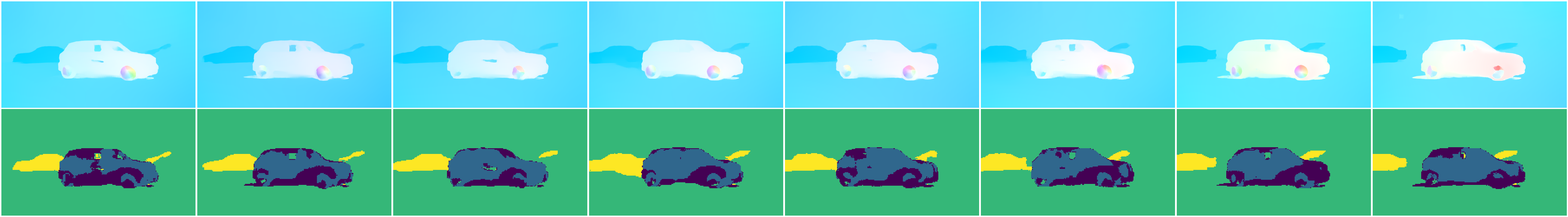}
\vspace{-0.5cm}
\caption{Illustration of the implicit temporal consistency ensured by our method with four masks. For each group, first row: input optical flow fields displayed with the HSV color code, second row: OFS maps. The color code is the same for all segmentation maps (dark blue: mask 1, light blue: mask 2, green: mask 3, yellow: mask 4). Examples are drawn from DAVIS2016 (from top to bottom): libby, dance-twirl, car-roundabout.}
\label{fig:temporal-coherency}
\vspace{-0.4cm}
\end{figure*}

{\color{black} Based on visual assessment of our results,} we also observed that our method implicitly ensures temporal consistency. Indeed, motion segments are consistently segmented over time. Moreover, they generally correspond to the same mask number within the same video. This {\color{black} seems to be}
particularly true for the background {\color{black} when looking at the segmentation results}. We illustrate this behaviour in Fig.\ref{fig:temporal-coherency}. This is an appealing property, since the output of the network could be easily exploiting for tracking or for higher-level dynamic scene understanding.

\vspace{-0.2cm}
\section{Discussion}
\label{discussion}

Our work takes its origin in the classical methods of image motion segmentation. The main idea is to group elements of similar nature under a common set of parameters. The novelty of CNN approaches is to train a model that extracts groups without relying on an iterative process at inference. In addition, they engage by design an implicit consistency making the network robust to corrupted observations.

As discussed in \cite{yang_unsupervised_2019}, region (or layer)-based motion segmentation approaches face the important challenge of specifying a proper model (or regularization) representing the flow within each segment. A too weak regularization may make the flow reconstruction trivial (i.e., the reconstruction is just a copy of the input flow). Conversely, a too strongly regularized model may make the flow reconstruction coarse, since it is likely to be not expressive enough.

Hereafter, we compare our design choices with the ones of the two competing unsupervised optical-flow segmentation methods: CIS \cite{yang_unsupervised_2019} and MoSeg \cite{yang_motion-grouping_2021}. 
They rely on a different view of the problem. CIS circumvents the regularization problem by inpainting the flow of the foreground region from the flow of the background one and vice versa. In this setting, any model, even unconstrained, could not correctly inpaint the flow in the event of a perfect segmentation, thus alleviating the problem of regularization.
MoSeg addresses the regularization problem by employing a representational bottleneck linked to the architectural structure of the network. Indeed, they force the flow to be reconstructed only from two latent representations of reduced size, one for the foreground and the other one for the background.

On our side, we address the regularization problem by using polynomial motion models. Of course, parametric motion models may lack expressiveness. However, we claim that it is a well-funded choice for the task of motion segmentation. Indeed, there are clear relationships between 2D parametric motion models in the image and 3D motion in the viewed scene \cite{PB-EF1993}. For instance, the 8-parameter quadratic motion model corresponds to the projection into the image of the rigid motion of a planar surface. Relying on those simple parametric motion models, our method achieves excellent performance as demonstrated in Table \ref{tab:results}.

Compared to CIS, our method bypasses the difficulty to train a complicated inpainting generative model. Compared to MoSeg, it is not attached to a dedicated architectural structure and enables light and efficient architectures for motion segmentation. 
In addition, the use of parametric motion models for flow representation and of a simple network structure makes the training easier. We advocate that this choice allows for a better generalisation. We consider this point as a critical feature in real-world applications and highlighted it in our evaluation protocol (Section \ref{eval}). 



Additionally, CIS and MoSeg methods are devoted to a two-mask configuration (moving object \textit{vs.} background), whereas our method can deal with multiple motions by design. Indeed, the use of parametric motion models allows us to perform multilayer motion segmentation (Fig.\ref{fig:4masks}, {\color{black}Table \ref{tab:multiple}), with for example the segmentation of the background moving objects and of articulated motions.} 



\vspace{-0.10cm}
\section{Conclusion}
\label{conclusion}

We have defined an original, real-time, unsupervised method for motion segmentation taking optical flow as input. We leveraged the EM paradigm to define a mathematically well-founded loss function and the training stage of our neural network. No manual annotation is required. We have also designed an effective and simple data augmentation scheme adapted to optical flow fields. In contrast to the classical EM algorithm, our method is not iterative at the inference stage 
and does not need to estimate parametric motion models at test time. In addition, our OFS method can handle by design the segmentation of multiple motions. Our method outperforms state-of-the-art comparable unsupervised methods on the DAVIS2016 and FBMS59 benchmarks, and is the second best for SegTrackV2 and MoCA, yielding the best overall performance. Additionally, the version with more than two masks opens a promising perspective on this point. Interestingly, our OFS method implicitly provides rather time-consistent segments.

Future work will further investigate the handling of the temporal dimension of the motion segmentation problem, {\color{black} the setting of the mask number in multi-mask segmentation and the mask selection for finding the independent moving objects}, and the use of our motion-related loss function as a plug-in regularizer for training networks devoted to other segmentation or interpretation tasks.

\onecolumn

\part*{Appendix}
\setcounter{section}{0}

\section{Loss Function}
Our loss function, given by eq.(12), is obtained as follows.
\begin{align*}
L(f, \theta, \phi) &= - ll(\theta, \phi) \ \mbox{using (11)}\\
&= -  \sum_i \sum_k  g_\phi(f)_i^k  \log p(f_i, z_i^k | \theta_k) + \sum_i \sum_k  g_\phi(f)_i^k \log g_\phi(f)_i^k \ \mbox{using (4)}\\
&= 
\begin{aligned}[t]
&-  \sum_i \sum_k  g_\phi(f)_i^k  \log p(f_i |  z_i^k , \theta_k) -  \sum_i \sum_k  g_\phi(f)_i^k  \log p( z_i^k) \  \mbox{using (9)}\\
&+ \sum_i \sum_k  g_\phi(f)_i^k \log g_\phi(f)_i^k\\
\end{aligned}\\
&=  
\begin{aligned}[t]
&\log Z \sum_i \sum_k  g_\phi(f)_i^k +  \frac{1}{\alpha}\sum_i \sum_k  g_\phi(f)_i^k  \delta(f_i, \theta^T_k \cdot c(i)) \ \mbox{using (10)}\\
&-  \sum_i \sum_k  g_\phi(f)_i^k  \log p( z_i^k) + \sum_i \sum_k  g_\phi(f)_i^k \log g_\phi(f)_i^k.
\end{aligned}
\end{align*}
As indicated in Section 4.2, in this work, we use a uniform prior for $p(z_i^k )$, i.e., $p(z_i^k )=\frac{1}{K}$. We can thus rewrite the loss $L$ as:
\begin{align*}
L(f, \theta, \phi) = &\log (K \ Z) \sum_i \sum_k  g_\phi(f)_i^k +  \frac{1}{\alpha}\sum_i \sum_k  g_\phi(f)_i^k  \delta(f_i, \theta^T_k \cdot c(i)) \\
&+ \sum_i \sum_k  g_\phi(f)_i^k \log g_\phi(f)_i^k.
\end{align*}
As $\sum_k  g_\phi(f)_i^k = 1, \ \forall i$, we have $\sum_i \sum_k  g_\phi(f)_i^k=I$, where $I$ is the number of sites $i$ in the frame. We finally obtain: 
\begin{align*}
L(f, \theta, \phi) =& \ I \ \log (K \ Z) +  \frac{1}{\alpha}\sum_i \sum_k  g_\phi(f)_i^k  \delta(f_i, \theta^T_k \cdot c(i))\\
&+ \sum_i \sum_k  g_\phi(f)_i^k \log g_\phi(f)_i^k.
\end{align*}

\section{Data Augmentation}

\newtheorem{theorem}{Theorem}
\theoremstyle{remark}
\newtheorem*{remark}{Remark}

\begin{theorem}\label{th:daug-inv}

Considering a function $ll$ such as: 

$$
ll(\theta, m, f) = \sum_i \sum_k  m_i^k  (\log p(f_i, z_i^k | \theta_k) - \log m_i^k),
$$
where $\theta  \in \R^{12}$ are the parameters of a parametric (quadratic) motion model, $m \in [0,1]^{W \times H \times K}$ is a segmentation mask and $f \in \R^{W \times H \times 2}$ is the input optical flow. We define $f_{\zeta} \in \R^{W \times H \times 2}$ as a parametric flow field generated using parameters $\theta_{\zeta}$ and positions $c(i)$ as $f_{\zeta, i} = \theta_{\zeta}\cdot c(i)$. We show that for all possible segmentation $m$ and parameters $\theta_{\zeta}$ we have : 

$$
\max_{\theta} ll(\theta, m, f) = \max_{\theta} ll(\theta, m, f + f_{\zeta}),
$$

\end{theorem}

\begin{figure*}
\label{fig:data-aug}
\includegraphics[width=\linewidth]{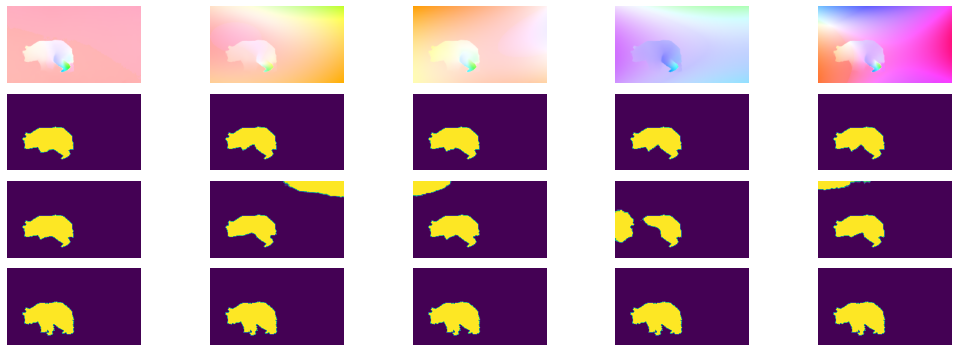}
\caption{Data augmentation adding a quadratic global motion on the optical flow field. From top to bottom : Optical flow field displayed with the usual HSV code \cite{middlebury}; Predicted mask with our network trained using data augmentation; Predicted mask with our network trained without data augmentation; Ground-truth segmentation mask.}
\end{figure*}

\begin{remark}
In this theorem, $ll$ and $c(i)$ are the lower bound and the polynomial terms, respectively, both of which are described in Sections 4.1 and 4.2 of the main paper. For the lower bound, we take $m$ instead of $g_{\phi}(f)$, since we are not specifically considering the network here. The goal here is to show that the lower bound we optimize in the paper is invariant to the added parametric global flow with respect to the segmentation. Thus, a segmentation will correspond to the same loss value, regardless of the perturbation of the input flow through the data augmentation, which encourages the network to produce global-motion-invariant segmentations. We can see an illustration of this property in Fig.\ref{fig:data-aug}. Furthermore, a direct consequence of this property is that the optimal $m^*$ segmentation is the same for both the original and augmented flows: 
$$
m^* = \arg \max_m [ \max_{\theta} ll(\theta, m, f)] = \arg \max_m [ \max_{\theta} ll(\theta, m, f+f_{\zeta})].
$$
\end{remark}

\vspace{0.2cm}

\begin{proof}

Starting from eq.4 and using eq.10 for the likelihood definition in Sections 4.1 and 4.2, we can write: 
\begin{align*}
ll(\theta, m, f) &= \sum_i \sum_k  m_i^k  (\log p(f_i, z_i^k | \theta_k) - \log m_i^k) \\
&= \sum_i \sum_k  m_i^k  \log p(f_i, z_i^k | \theta_k) - m_i^k  \log m_i^k \\
&= \sum_i \sum_k  m_i^k  \log p(f_i | z_i^k, \theta_k) + m_i^k \log \bigg(\frac {p(z_i^k)}{p(m_i^k)}\bigg)\\
&= - \sum_i \sum_k  m_i^k \delta(f_i, \theta_k^T \cdot c(i))  + \kappa,
\end{align*}
where $\kappa \triangleq m_i^k \log \bigg(\frac {p(z_i^k)}{p(m_i^k)Z} \bigg)$ is independent of $f$ and $\theta$. See Theorem \ref{th:z-fac} justifying this statement for the normalisation factor $Z$. Consequently, we have: 
$$
ll(\theta, m, f+f_{\zeta}) = - \sum_i \sum_k  m_i^k \delta(f_i + f_{\zeta,i}, \theta_k^T \cdot c(i))  + \kappa.
$$
Using translation invariance of our distance function $\delta$ and the additivity property of our parametric motion models (that are linear with respect to the parameters), we can write: 
\begin{align*}
\delta(f_i + f_{\zeta,i}, \theta_k^T \cdot c(i)) &= \delta(f_i, \theta_k^T \cdot c(i) - f_{\zeta,i}) = \delta(f_i, \theta_k^T \cdot c(i) - \theta_{\zeta}^T \cdot c(i)) \\&= \delta(f_i, (\theta_k - \theta_{\zeta})^T \cdot c(i))  = \delta(f_i, \tilde \theta_{k}^T \cdot c(i)).
\end{align*}
Thus, we have: 
$$
\max_{\tilde  \theta} ll(\tilde  \theta, m, f+f_{\zeta}) = \max_{\tilde  \theta} - \sum_i \sum_k  m_i^k \delta(f_i, \tilde \theta_{k}^T \cdot c(i)) + \kappa.
$$
With a change of variable, we get: 
$$
\max_{\theta} ll(\theta, m, f) = \max_{\theta} ll(\theta, m, f +f_{\zeta}).
$$

\end{proof}

\section{Normalisation factor}

\begin{theorem}\label{th:z-fac}
Let a translationally invariant function $\delta$ : $\mathbb{R}^{D} \times \mathbb{R}^D \mapsto \mathbb{R} $ such that, $\forall\ a, e, c \in \mathbb{R}^{D} \times \mathbb{R}^D, \ \delta(a+c, e+c) = \delta(a,e)$, and a probability density defined as: 
$$p(y |x; \theta) = \frac1Z \exp(-\delta(y, \theta^Tx)).$$

Then, the normalisation factor $Z$ is independent of $\theta$.
\end{theorem}

\vspace{0.5cm}
\noindent In particular, this result is true for the $p$-norm defined by: $$\delta(a, b) = ||a-b||_p = \big(\sum_i |a_i-b_i|^p\big)^{1/p}.$$

\begin{proof} 
The normalisation factor is defined by: $$Z \triangleq \int_\mathbb{R^D} \exp(-\delta(y, \theta^Tx))\mathrm{d}y.$$

Taking $b \triangleq y - \theta^Tx$, we get:

$$Z = \int_\mathbb{R^D} \exp(-\delta(b+\theta^Tx, \theta^Tx))\mathrm{d}b = \int_{\mathbb{R^D}} \exp(-\delta(b,0))\mathrm{d}b.$$

Thus, $Z$ is independent of $\theta$, $x$ and $y$, and only depends on function $\delta$. \\

For the $p$-norm: $$\delta(a+c, e+c) = ||a+c -e -c||_p = ||a - e||_p = \delta(a,e).$$
\end{proof}

\section{Further analysis of the classical EM}
\label{classicEM}

\begin{figure}[t!]
\includegraphics[width=\linewidth]{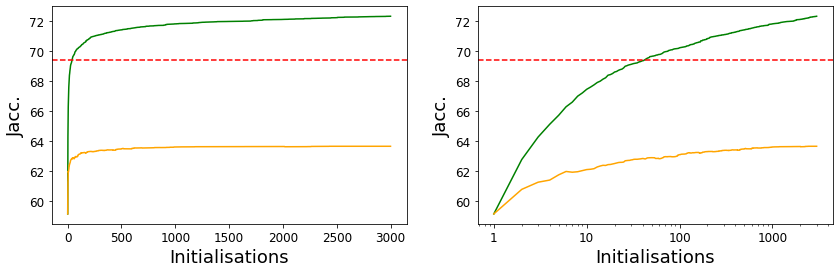}
\caption{Sensitivity with respect to the initialization of the classical EM algorithm in contrast to our trained model. First, we plot the evolution of the (speculative) optimal Jaccard index of the classical EM algorithm of Section 3 of the main text (green curve) with respect to the number of available random initialisations of the parametric motion parameters $\theta$ on the DAVIS2016 validation set. For all videos, the optimal (speculative) Jaccard index is obtained by selecting the best segmentation for each input flow among all available initialisations. Secondly, the yellow curve plots the evolution of the Jaccard index of the classical EM algorithm over the available initialisations, when selecting the initialisation with the highest likelihood, the only means available in practice without ground-truth. The dashed red horizontal line corresponds to the value of the Jaccard index obtained with our network on the same validation set. The dashed blue vertical line indicates when the optimal Jaccard index of the classical EM algorithm exceeds the score of our network. The figure on the right involves a logarithmic scale to ease the visualisation of the first part of the graph.}
\label{fig:classical_EM}
\vspace{-0.2cm}
\end{figure}

In this subsection, we elaborate on the iterative parametric motion segmentation based on the classical EM algorithm, described in Section 3 of the main text. More precisely, we study the influence of the initialisation of $\theta$, parameters of the motion models, on the classical EM algorithm performance. We perform this experiment still using quadratic motion models. In doing so, we also highlight the impact of the neural network alone on the performance of our method.


Even if EM algorithm is guaranteed to converge to a local optimum \cite{murphy_book2012}, the quality of the optimum reached is highly dependent on the initialisation of $\theta$. We take as conditional likelihood $p(f_i, z_i^k | \theta_k)$ the Gaussian distribution, which leads to the squared $L_2$ once applying the log function. We evaluate EM on the DAVIS2016 benchmark. Let us recall that the performance criterion is the Jaccard index computed on the primary moving object, not the value of the likelihood being maximised.

To emphasize the dependence of EM on the initialization of $\theta$, we study the evolution of the Jaccard index over the validation set with respect to the number of available random initializations. For each point of the curve, we try available initialisations, and select for each frame the one that maximizes the likelihood. We repeat a second time this experiment, but now using the Jaccard index as selection criterion. Of course, this optimal Jaccard index remains speculative in practice, since we need the ground truth to find the best Jaccard score for each sample. Let us stress that the computational cost involved is enormous in both cases.

We summarize the results obtained on the DAVIS2016 validation set in Fig.\ref{fig:classical_EM}. We observe that the optimal Jaccard index for the classical EM algorithm increases with the number of initialisations until reaching a plateau where testing additional initialisation does not improve the results. To ensure this behaviour, we extended this experiment up to a huge number of 3000 initialisations. 
The optimal (speculative) Jaccard index for EM goes up to $72.3$. The Jaccard index for EM when selecting the initialisation with the maximum likelihood, the only means available in practice without ground-truth, tops out at $63.7$ .
This demonstrates the impact of the initialisation step on the performance of the classical EM algorithm. As mentioned above, the optimal Jaccard index is only speculative. Thus, we consider this score of $72.3$ as a gold standard.
%


Importantly, Fig.\ref{fig:classical_EM} shows that the performance of our network is relatively close to the gold standard given by the speculative optimal Jaccard index of the classical EM algorithm. Besides, the latter needs a number of 41 initialisations to exceed the score of our network. This is already a large number, since, even with our efficient GPU implementation, it takes $25$ minutes to run the 41 initialisations per sample over the DAVIS2016 validation set on a Tesla P100.

In Fig.\ref{fig:classical_EM}, we also observe that our network outperforms by a large margin the classical EM algorithm when the initializations are selected according to the highest likelihood, even for many tested initialisations, the only means available in practice without ground-truth. In addition, the network does not need motion models, and of course, does not involve any initialisation on motion parameters $\theta$, when segmenting the optical flow.
Moreover, we can easily adopt robust loss functions for the network training instead of the squared $L_2$ loss.

In Table \ref{tab:compare-classic}, we report a quantitative comparison based on several aspects between our method and the classical EM. Accuracy and inference time were computed on the DAVIS2016 validation dataset. As we mentioned above, results using the classical EM strongly depend on the initialisation of the motion parameters. To assess the accuracy of the classical EM, we report first the average score obtained using randomly chosen motion parameters, and secondly, the score obtained by selecting the best initialization from a set of 500 random initialisations. We used 500 here since this is the number of initialisations where the Jaccard score starts to plateau at its optimal value, as shown in Fig.\ref{fig:classical_EM}. By design, our method involves a segmentation network with a very large number of parameters to train and store. However, as demonstrated in Table \ref{tab:compare-classic}, our method exhibits strong advantages compared to the classical EM regarding inference time and outperforms it by a large margin in term of accuracy.

\begin{table}[]
    \resizebox{\columnwidth}{!}{
	\begin{tabular}{|c||c|c|c|}\hline
		&  \multicolumn{2}{c|}{Classical EM} & Our network \\ \hline
		Number of inferences & 1 & 500 &1                          \\ \hline
		Average Jaccard (DAVIS2016) & 59.0 $\pm$ 0.35 & 63.5  & 69.3    \\ \hline
		Inference time / frame (in ms) & 9.6 &  4800 & 8       \\ \hline
		Number of parameters  &\multicolumn{2}{c|}{W $\times$ H + 12 $\times$ K} & 497 millions + 66 $\times$ K  \\ \hline
		Nb. param. (for K=2,W=128,H=224) & \multicolumn{2}{c|}{28696} & 497 millions \\ \hline
	\end{tabular}}
	\caption{Quantitative comparaison between classical EM and our trained network on several aspects. Inference time is computed on Tesla-V100 in all cases. Accuracy (average Jaccard) is evaluated on DAVIS2016 validation set. The size of the input flow is noted (W,H) and the number of layers K. Figures for the classical EM are given for two configurations: 1) just applying it once for a random initialisation, but for the accuracy score and a fair comparison, we prefer to give the mean and standard deviation of the average Jaccard index estimated over 3000 initialisations; 2) selecting the best initialisation on every frame among 500 initialisations.}
	\label{tab:compare-classic}
\end{table}
\label{tab:compare_em}


\section{Algorithm}

\begin{figure}[H]
\centering
{\footnotesize
\begin{minted}[linenos]{python}
def TrainingStep(Flow, ConvNet, alpha=0.01, learning_rate=0.01) :
    Segmentation = SoftMax(ConvNet(Flow)) # Extract Segmentation Probabilistic Mask using any Convolutional Net
    Theta = ComputeThetaOptim(Flow, Segmentation) # Compute theta for each segment minimizing ll
    Theta = Theta.detach() # Stop Gradient on theta estimation ( alternate optimisation )
    
    ll = CoherenceLoss(Theta) + alpha*Entropy(Segmentation) # Equation in Section 4.2
    ConvNet = OneStepOptimiseNetwork(ll, ConvNet, learning_rate) # One step of Adam on network's weights.

def ComputeThetaOptim(Flow, Segmentation, delta='L1_norm') : 
    for k in {1..K} : # For each motion segment k
    	ll_k = Segmentation_k*delta(Flow, Theta_k . c(i)) # Coherence loss associate to the segment
        Theta_k = Minimize(ll_k) # LBGFS to minimize ll_k
    return Theta

def InferenceStep(Flow, ConvNet) : 
    return ArgMax(ConvNet(Flow)) # For each site return the most likely mask
\end{minted}
}
	\vspace*{-1em}
	\caption{Pseudo-Code of the training and inference steps of our method. Let us emphasize that each training step and the inference step are not iterative.}
	\label{pytorch-code}
	\vspace*{-1em}
\end{figure}

\begin{figure*}[t!]
\label{failure-cases}
\includegraphics[width=\linewidth]{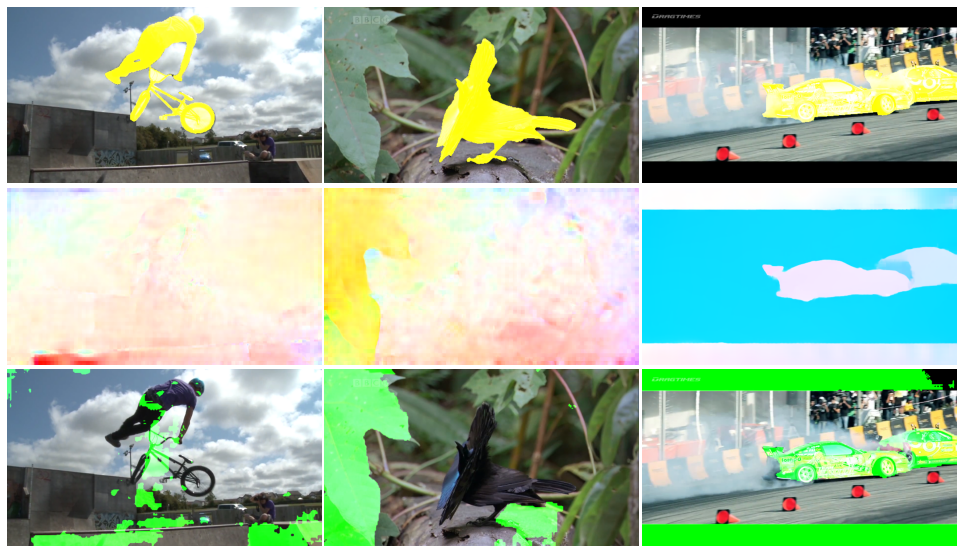}
\caption{Examples of motion segmentation results obtained with our method with two masks, on the videos bmx, birds of paradise and drift of the SegTrackV2 dataset. First row: a frame of the video with the ground-truth superimposed in yellow. Second row: the input flow field displayed with the HSV color code. Third row: the segmentation produced by our method superimposed in green on the corresponding image.}
\end{figure*}

\noindent Code available at : https://github.com/Etienne-Meunier/EM-Flow-Segmentation

\section{Failure cases on SegTrackV2}
\label{sec:multi_seg}

We observed that the optical flow is not well estimated by the RAFT method in several frames of SegTrackV2 dataset, causing our algorithm to fail on those frames. These failure cases are reported in Fig.\ref{failure-cases}.

\section{Detailed results per sequences of the datasets}
Hereafter, we report detailed results through tables collecting the evaluation scores obtained by our method for every sequence of the four datasets, DAVIS2016, SegTrackV2, FBMS59 and MoCA.

\subsection{DAVIS2016}
\csvnames{davisnames}{1=\Sequence,2=\JM,3=\JO,4=\JD,5=\FM, 6=\FO, 7=\FD}
\csvstyle{Rt}{tabular=|l|c|c|c|c|c|c|,table head=\hline Sequence & $\mathcal{J}$ (M) & $\mathcal{J}$ (O) & $\mathcal{J}$ (D) & $\mathcal{F}$ (M) & $\mathcal{F}$ (O) & $\mathcal{F}$ (D) \\\hline\hline,late after line=\\\hline,davisnames}

\begin{table}[H]\centering
\begin{filecontents*}{csv/1tdcjfqp_davisval.csv}
      Sequence      ,  J(M)  ,  J(O)  ,  J(D)  ,  F(M)  ,  F(O)  ,  F(D)  
     blackswan      ,0.499,0.5,-0.091,0.608,1,-0.022
     bmx-trees      ,0.602,0.782,0.241,0.789,0.923,0.142
     breakdance     ,0.754,0.939,-0.013,0.789,1,0.012
       camel        ,0.827,1,0.139,0.793,1,0.146
   car-roundabout   ,0.908,1,-0.035,0.804,1,-0.067
     car-shadow     ,0.89,1,0.042,0.83,1,0.044
        cows        ,0.856,1,0.031,0.773,1,0.032
    dance-twirl     ,0.8,1,-0.087,0.831,1,-0.024
        dog         ,0.808,1,-0.09,0.743,0.983,-0.085
   drift-chicane    ,0.674,0.78,-0.193,0.755,0.86,-0.065
   drift-straight   ,0.896,1,0.04,0.855,1,0.217
        goat        ,0.134,0,0.099,0.408,0.25,-0.049
   horsejump-high   ,0.805,1,0.071,0.873,1,-0.001
     kite-surf      ,0.427,0.271,0.219,0.483,0.396,0.004
       libby        ,0.461,0.553,0.47,0.665,0.766,0.297
   motocross-jump   ,0.651,0.737,0.037,0.576,0.658,0.093
 paragliding-launch ,0.623,0.667,0.308,0.327,0.205,0.406
      parkour       ,0.563,0.551,-0.014,0.701,0.847,0.118
   scooter-black    ,0.784,1,-0.262,0.669,1,-0.099
      soapbox       ,0.891,1,0.015,0.865,1,0.008
      Average       ,0.693,0.789,0.046,0.707,0.844,0.055
\end{filecontents*}
\csvreader[Rt]{csv/1tdcjfqp_davisval.csv}{}{\Sequence & \JM & \JO & \JD & \FM & \FO & \FD }
\caption{Results given for every sequence of DAVIS2016 dataset. Reported score is the average Jaccard score over frames in the sequence. Last row is the average over sequences scores.
$\mathcal{J}$ is the Jaccard index and $\mathcal{F}$ is the Countour Accuracy. The Mean ($M$) is the average of the score, the Recall ($O$) is the fraction of frames with a score higher than $0.5$ and the Decay ($D$) is the degradation of the score over time in the sequence. More details in \cite{pont-tuset_benchmark_2016}.}
\end{table}


\subsection{SegTrackV2}

\csvstyle{Rt}{tabular=|l|c|,
table head=\hline Sequence & Jacc ($\mathcal{J}$)\\\hline\hline,
late after line=\\\hline,
head to column names}

\begin{table}[H]\centering
\begin{filecontents*}{csv/1tdcjfqp_segtrack.csv}
Sequence,Score
bird of paradise,54.4
birdfall,40.7
bmx,69.7
cheetah,39.7
drift,36.1
frog,77.6
girl,64.0
hummingbird,65.5
monkey,62.0
monkeydog,12.6
parachute,92.7
penguin,32.6
soldier,72.9
worm,41.8
Seq. Avg.,54.5
Frames. Avg.,55.5
\end{filecontents*}
\csvreader[Rt]{csv/1tdcjfqp_segtrack.csv}{}{\Sequence & \Score}
\caption{Results given for every sequence of SegTrackV2 dataset. Reported score is the average Jaccard score over annotated frames in the sequence.}
\end{table}

\subsection{FBMS59}

\begin{table}[H]\centering
\begin{filecontents*}{csv/1tdcjfqp_fbms.csv}
Sequence,Score
camel01,65.0
cars1,87.1
cars10,33.2
cars4,86.6
cars5,83.7
cats01,68.7
cats03,78.3
cats06,39.4
dogs01,72.9
dogs02,69.8
farm01,81.0
giraffes01,34.2
goats01,43.7
horses02,75.9
horses04,67.8
horses05,43.0
lion01,56.1
marple12,50.0
marple2,70.4
marple4,80.6
marple6,35.5
marple7,62.4
marple9,31.0
people03,53.3
people1,78.7
people2,86.1
rabbits02,44.6
rabbits03,40.2
rabbits04,42.9
tennis,72.3
Seq. Avg.,61.1
Frames. Avg.,57.8
\end{filecontents*}
\csvreader[Rt]{csv/1tdcjfqp_fbms.csv}{}{\Sequence & \Score}
\caption{Results given for every sequence of FBMS59 dataset. Reported score is the average Jaccard score over annotated frames in the sequence.}
\end{table}

\subsection{MoCA}

\csvstyle{Rtl}{longtable=|l|c|,
table head=\caption{Results given for every sequence of MoCA dataset. Reported score is the average Jaccard score over frames in the sequence computed using bounding box annotation as described in Section 5.2. of the main text.}\\\hline Sequence & Jacc ($\mathcal{J}$)\\\hline\hline,
late after line=\\\hline,
head to column names}

\begin{filecontents*}{csv/1tdcjfqp_moca.csv}
Sequence,Score
arabian horn viper,71.0
arctic fox,37.8
arctic fox 1,85.5
arctic wolf 0,63.4
arctic wolf 1,50.8
bear,61.3
black cat 0,53.5
black cat 1,8.6
crab,68.1
crab 1,29.3
cuttlefish 0,23.5
cuttlefish 1,20.0
cuttlefish 4,64.5
cuttlefish 5,72.1
dead leaf butterfly 1,75.7
desert fox,27.0
devil scorpionfish,90.3
devil scorpionfish 1,92.5
devil scorpionfish 2,85.7
egyptian nightjar,72.4
elephant,80.4
flatfish 0,59.9
flatfish 1,64.8
flatfish 2,82.1
flatfish 4,22.1
flounder,89.3
flounder 3,21.4
flounder 4,77.7
flounder 5,69.7
flounder 6,80.9
flounder 7,51.5
flounder 8,69.7
flounder 9,71.6
fossa,16.6
goat 0,66.6
goat 1,72.2
groundhog,56.5
hedgehog 0,43.4
hedgehog 1,55.6
hedgehog 2,69.9
hedgehog 3,50.9
hermit crab,68.8
ibex,25.8
jerboa,54.8
jerboa 1,41.7
lichen katydid,67.8
lion cub 0,80.1
lion cub 1,54.2
lion cub 3,23.0
lioness,25.9
marine iguana,44.8
markhor,80.0
meerkat,88.6
mountain goat,73.9
nile monitor 1,38.6
octopus,63.6
octopus 1,55.4
peacock flounder 0,94.6
peacock flounder 1,86.1
peacock flounder 2,89.4
polar bear 0,0.0
polar bear 1,21.7
polar bear 2,75.5
pygmy seahorse 2,47.0
pygmy seahorse 4,68.4
rodent x,63.4
scorpionfish 0,67.2
scorpionfish 1,63.7
scorpionfish 2,85.9
scorpionfish 3,82.8
scorpionfish 4,84.1
scorpionfish 5,85.7
seal 1,86.2
seal 2,56.2
seal 3,51.3
shrimp,69.4
snow leopard 0,75.6
snow leopard 1,82.7
snow leopard 2,90.9
snow leopard 3,69.7
snow leopard 6,87.8
snow leopard 7,68.1
snow leopard 8,61.3
snowy owl 0,59.8
spider tailed horned viper 0,40.7
spider tailed horned viper 1,52.5
spider tailed horned viper 2,81.4
spider tailed horned viper 3,83.3
Seq. Avg.,61.9
Frames. Avg.,61.8
\end{filecontents*}
\csvreader[Rtl]{csv/1tdcjfqp_moca.csv}{}{\Sequence & \Score}


\end{document}